\documentclass{article}
\usepackage{ijcai25}

\usepackage{times}
\usepackage{soul}
\usepackage{url}
\usepackage[hidelinks]{hyperref}
\usepackage[utf8]{inputenc}
\usepackage[small]{caption}
\usepackage{graphicx}
\usepackage{amsmath}
\usepackage{amsthm}
\usepackage{booktabs}
\usepackage{algorithm}
\usepackage{algorithmic}
\usepackage[switch]{lineno}

\usepackage{listings}
\title{Large Language Models as Common-Sense Heuristics}

\author{
Andrey Borro
\and
Patricia J Riddle\and
Michael W Barley\And
Michael J Witbrock\\
\affiliations
University of Auckland
\emails
abor198@aucklanduni.ac.nz,
\{p.riddle, m.barley, m.witbrock\}@auckland.ac.nz
}

\begin{document}

\maketitle

\begin{abstract}
    While systems designed for solving planning tasks vastly outperform Large Language Models (LLMs) in this domain, they usually discard the rich semantic information embedded within task descriptions. In contrast, LLMs possess parametrised knowledge across a wide range of topics, enabling them to leverage the natural language descriptions of planning tasks in their solutions. However, current research in this direction faces challenges in generating correct and executable plans. Furthermore, these approaches depend on the LLM to output solutions in an intermediate language, which must be translated into the representation language of the planning task. We introduce a novel planning method, which leverages the parametrised knowledge of LLMs by using their output as a heuristic for Hill-Climbing Search. This approach is further enhanced by prompting the LLM to generate a solution estimate to guide the search. Our method outperforms the task success rate of similar systems within a common household environment by 22 percentage points, with consistently executable plans. All actions are encoded in their original representation, demonstrating that strong results can be achieved without an intermediate language, thus eliminating the need for a translation step.

\end{abstract}

\section{Introduction}
Emergent reasoning capabilities in Large Language Models (LLMs) have prompted research into their ability to solve planning tasks and generate plans in the formal languages used to encode planning domains~\cite{rao:llm-modulo,hao:translate-planning,silver:genplanning}. This research has demonstrated that LLMs have difficulty with solving even simple tasks and that recent progress, with the introduction of Large Reasoning Models (LRMs)~\cite{openai:o1-systemcard}, is insufficient to deal with common planning challenges, such as solving obfuscated domains or analysing the solvability of a given task. Furthermore, initial LRM success in solving small planning tasks fails to scale past toy problems~\cite{rao-o1}. 

The trend in improvement shows that even a modest increase in planning ability comes at the price of a larger, hence, computationally and financially costlier, LLM. This raises questions about the practicality of LLM-based planning, especially when compared to traditional planning systems such as FastDownward~\cite{helmert:fastdown}, which vastly outperform modern LLMs on planning tasks for a fraction of cost and time. 

One advantage that LLMs have over traditional planning systems is the models' ability to store vast amounts of parametrised knowledge about a wide variety of topics and interact with the semantic information in their input text. Most planning systems make no distinction between the labels \textit{key} and \textit{object1}, but an LLM is likely to infer that a key could be used to unlock a door or a chest. 

Research in utilising LLMs for planning in everyday environments, such as a common household, takes advantage of the parametrised world-knowledge of an LLM through fine-tuning existing models~\cite{finetunegpt2} or through directly evaluating the pre-trained performance of LLMs on generating plans from task descriptions~\cite{zeroshot2022}. Both methods provide the LLM with a high-level task description (e.g., `slice an apple') and expect a sequence of actions in return. A common challenge for this research comes from ensuring the executability of these action sequences, which require translation from a high-level language to a low-level language used by planning systems to operate within finite, discrete action spaces.

Recent studies demonstrate that increasing the amount of information provided to the LLM alongside the task description can improve both success rates and executability. Hao et al.~\shortcite{worldmodel2023} achieve this by providing an explicit world model, using a natural-language description of the current state, as part of the input to an LLM solving simple block manipulation tasks. 

Other research proposes to rank planning actions at each step based on a combination of the LLM’s sampling probability and an affordance function; that is, a function which maps actions to a probability estimate of their executability~\cite{saycan2022}. An affordance function can be implemented by using the Q-value function from a pre-trained model. Additionally, many other implementations are possible and can even be applied at the token level~\cite{huang:decoding}.

Most similar to our research, ProgPrompt~\cite{progprompt2023} has shown improved rates of success and executability through generating and then iteratively refining solutions in a low-level programming language, increasing the accuracy of the translation step. This result challenges the need for the LLM to output its solution in a high-level language. However, it falls short of removing the translation step altogether. 

We propose a method to utilise the world knowledge of an LLM by tasking it with action selection for a local search algorithm which controls an agent operating within a household environment. This allows it to function as a heuristic for the search, replacing the numeric heuristics that are more conventionally used. We extend our method by proposing a secondary heuristic in the form of an initial solution estimate to aid the LLM in pursuing long-term goals while selecting actions. 

Our results demonstrate that our approach is capable of generating plans to perform simple tasks in a household environment, with a success rate 22 percentage points higher than that of ProgPrompt. Furthermore, our method consistently generates executable plans, even when the tasks are solved incorrectly. The performance of our system shows that intermediate languages, whether high-level or low-level, are not necessary and give no notable benefit over operating directly in the same representation language as the agent being controlled by the LLM. 

\section{Background}
\paragraph{Automated Planning.}
\label{planning_defintion}
Automated Planning is a branch of Artificial Intelligence primarily focused on the generation of plans. Plans are sequences of actions within a given environment that can be sequentially applied to an initial state. 

A planning task $\Pi$ is a 6-tuple $(S, s_i, S_g, A, f_a, f_t)$ consisting of: 
\begin{itemize}
    \item A finite and discrete set of all states $S$, also known as the state space.
    \item The initial state $s_i \in S$. 
    \item The set of goal states $S_g \subseteq S$. In our tasks, the set of goal states is defined by a set of goal conditions for each task, which outline the criteria for a state to be a goal state. 
    \item The set of actions $A$.
    \item The applicability function $f_a$, which returns $f_a(s) \subseteq A$, i.e. the set of all actions applicable at state $s$. In our paper, the applicability function is defined by the preconditions of our action schema, discussed in Section~\ref{section:vhome}  
    \item The transition function $f_t$, which maps a state $s$ and applicable action $a$ to the resultant state $s'$ of applying $a$ to $s$, such that $s' = f_t(s, a)$, $a \in f_a(s)$. As above, the transition function is defined by the effects of our action schema.   
\end{itemize}

In practice, most planning systems~\cite{planning1970,idastar2021} perform graph search on a graph $G = (V,\, E)$, where $V$ is the set of vertices and $E$ is the set of edges. Vertices are states $s \in S$, while edges are formed from the transition function $f_t$, where 
\[
\forall\, u, v \in V, \quad (u, v) \in E \iff \exists\, a \in A \colon f_t(u, a) = v.
\]

A solution for a given planning task is a plan $a_1 ... a_n$ such that one can form a path $v_0 ... v_{n}$ through $G$, where $v_i = f_t(v_{i-1},\,a_i)$, $v_0 = s_i$ and $v_n = s_g \in S_g$; that is, a plan that transforms $s_i$ into $s_g$ when applied sequentially. Heuristics~\cite{heuristic1968,heuristics1982} are often used to guide the search and reduce the number of vertices that must be visited for a solution to be found. 

Global search algorithms, such as A$^*$~\cite{aima2016} or BFS~\cite{aima2016}, store a `frontier' of vertices in memory, allowing the search to backtrack or switch to a different path through the graph entirely. These search algorithms usually have guarantees on finding a path from the initial state to a goal state if one exists (completeness), and sometimes have guarantees on that path being minimal cost (optimality), depending on other factors. 

Local search algorithms~\cite{local1999,local2002}, such as Hill-Climbing~\cite{local2011}, only store the current state and transform it by applying the transition function `in-place'. The action is usually chosen by minimising or maximising across some heuristic function. These algorithms typically have no guarantees of completeness or optimality and can have difficulty with getting stuck in local minima or maxima. Our research forgoes a numeric heuristic in favour of allowing the LLM to choose the action to take at a given state. 

\paragraph{Large Language Models.}
Large Language Models (LLMs)~\cite{llmcodex2021,llmeval2024,llmmultimodal2023} are very large neural networks, typically containing billions of parameters. These models are trained through unsupervised learning on vast corpora of human-written text. Prominent large language models, such as GPT-3~\cite{gpt3} and GPT-4~\cite{gpt4}, leverage the self-attention mechanism of the Transformer~\cite{attention2017} architecture to effectively process long text sequences in parallel and capture non-contiguous word dependencies.


Prompt engineering~\cite{prompteng-patterncatalog2023,prompteng-promptcatalog2024,shin:autoprompt} is a common approach to enhancing the reasoning performance of LLMs without fine-tuning their parameters or altering their architecture. With prompt engineering, specific prompts are crafted to influence how the model's parametrised knowledge interacts with the input text. This can involve techniques like providing worked examples, phrasing the query in a question-and-answer format, or asking the model to `think through its answer step-by-step'~\cite{kojima:step-by-step}.


LLMs demonstrate an ability to infer, generalise, and apply learned patterns to various scenarios~\cite{reason2021,reason2022}. However, there is research that claims that LLMs show little evidence of true emergent reasoning capabilities~\cite{emergent2023} and have difficulty with more complex forms of reasoning~\cite{gael2023}. Similarly, there are questions about the capacity of LLMs to compete with traditional planning systems~\cite{cantplan2024}.

\section{Methodology}
A comprehensive set of all prompts, environments and experiment data, as well as the source code to replicate our experiments and important documentation for VirtualHome, can be found at the link in Section~\ref{section:declaration}.

\subsection{VirtualHome Environment}
\label{section:vhome} 
VirtualHome~\cite{virtualhome2018} is a simulation environment designed to model a household in 3D space. It provides an interactive platform to test and evaluate AI agents in terms of their ability to perform simple household tasks like cleaning, cooking, or fetching items. The environment uses its own internal language to represent finite, discrete, and deterministic actions. An agent can execute sequences of actions to interact with the environment.

VirtualHome actions are comprised of an action command and an ordered list of associated objects. The length of the list is dependent on the action command in question. For example, \texttt{STANDUP} has no associated objects, \texttt{PICKUP} has one and \texttt{PLACEON} has two. Objects in VirtualHome have both names and IDs to avoid ambiguity around identically named objects. 

The VirtualHome language represents actions in the following format:
\[\texttt{[VERB]<object$_1$>(id$_1$)...<object$_n$>(id$_n$)}\]
For example, the action to put salmon in the microwave is expressed as\[\texttt{[PUTIN]<salmon>(311)<microwave>(297)}\] 
This low-level language is used by the VirtualHome environment to represent, understand and execute actions. 

As most of the functionality of the VirtualHome Environment is not required for our experiments, this paper instead uses a planning domain crafted from its documentation and operating in the VirtualHome language. The action schemas for our domain are taken directly from the VirtualHome documentation. Action schemas are blueprints for actions, describing their structure, preconditions and effects. They become actions when grounded by objects from the environment.

For example, the action schema 
\[\texttt{[PUTIN]<object$_1$>(id$_1$)<object$_2$>(id$_2$)}\] 
becomes the action 
\[\texttt{[PUTIN]<salmon>(311)<microwave>(297)}\] 
when grounded by the objects 
\[\texttt{salmon(311), microwave(297)}\] 

For the purpose of this paper, object IDs are not taken into account by the agent, though they are still stored internally. As such, for the rest of this paper, we will also omit the IDs when discussing objects or actions.

The initial states for all tasks in this paper are derived from environment graphs used by the VirtualHome Environment, which have a one-to-one mapping with states within our environment. The environment graphs used in our paper contain around 450 distinct objects with over 14 thousand distinct relationships.

We extend the VirtualHome environment by providing additional functionality, such as applying
\[\texttt{in(microwave, \textit{object})} \rightarrow \texttt{heated(\textit{object})}\]
to all objects in the state when \texttt{on(microwave)} becomes true; that is, anything inside a microwave becomes heated when it turns on. These extensions are made to allow the testing of more complex tasks without significant changes to the predicates or action schema. 

\subsection{Overview}
\begin{figure*}[t]
    \centering
    \includegraphics[width=\linewidth]{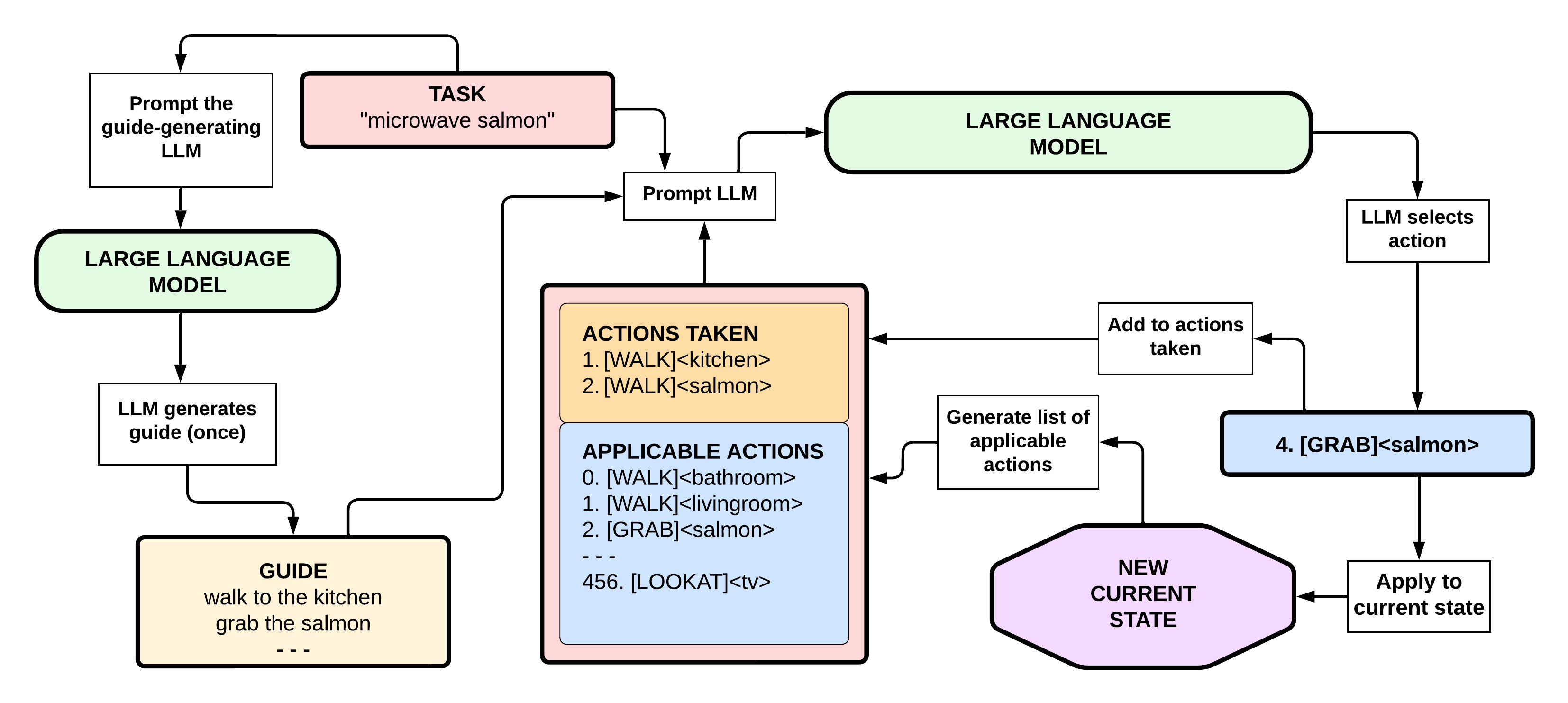}
    \caption{A diagram of our approach. The action selection process is shown in a simplified form, and error correction is omitted. A detailed description of both can be found in Section~\ref{section:Overview}}
    \label{fig:diagram}
\end{figure*}

Figure~\ref{fig:diagram} presents a diagram of our proposed approach.
\label{section:Overview}

\paragraph{Task Descriptions.}
The agent receives a high-level task description, such as `microwave salmon'. Each task description has an associated pair of human-written goal and failure conditions. For example, `microwave salmon' has goal conditions 
\[\texttt{heated(salmon) $\land$ on(microwave)}\] 
and failure conditions 
\[\texttt{heated(salmon) $\land$ $\neg\,$on(microwave)}\]

This means that the task is successfully completed when the salmon is heated, and the microwave is on, and failed if the salmon is heated but the microwave is off; that is, it was heated elsewhere. The agent has no access to the goal and failure conditions, but the system simulating the environment is able to check whether the conditions are satisfied and inform the agent of success or failure accordingly.

\begin{table*}[t]
\centering
\begin{tabular}{*{2}{l}}
\toprule
\textbf{High-Level Guide} & \textbf{Low-Level Guide} \\ 
\midrule
1. Walk to the kitchen & 1. walk $\mid$ kitchen \\ 
2. Pick up the coffeepot & 2. grab $\mid$ coffeepot \\ 
3. Walk to the living room & 3. walk $\mid$ livingroom \\ 
4. Put the coffeepot on the coffeetable & 4. placeon $\mid$ coffeepot $\mid$ coffeetable \\ 
5. Walk over to the kitchen & 5. walk $\mid$ kitchen \\ 
6. Get the cupcake & 6. grab $\mid$ cupcake \\ 
7. Go to the living room & 7. walk $\mid$ livingroom \\ 
8. Place the cupcake on the coffeetable & 8. placeon $\mid$ cupcake $\mid$ coffeetable \\  
\bottomrule
\end{tabular}
\caption{A high-level guide and a low-level guide for the task \textit{`bring coffeepot and cupcake to the coffee table'}}
\label{tab:guide_example}
\end{table*}

\paragraph{Guides.}
In Huang et al.~\shortcite{zeroshot2022}, an LLM is used to generate a high-level solution directly from a high-level task description. Rather than generating and executing such a solution directly, we propose to use it as a guide to aid an LLM in selecting actions during local search. 


The paper explores two types of guides. High-Level Guides are represented in English like the action plans generated by Huang et al.~\shortcite{zeroshot2022}. Low-Level Guides are also solution estimates for the planning task but are represented in the VirtualHome language. An example of the two types of guide for `put the coffeepot and the cupcake on the coffee table' is presented in Table~\ref{tab:guide_example}.  

\paragraph{Prompt Structure.}
Information about the agent is provided to the LLM through a constructed prompt. The prompt is composed of a high-level task description, a guide, a list of actions that the agent has already taken, and an indexed list of actions to choose from. A small hint about the environment, namely the necessity to walk to any given object before interacting with it, is also included in the prompt. All actions are represented in the VirtualHome language.

\paragraph{Action Selection.}
It was a concern that LLMs, particularly smaller and cheaper models, could struggle with keeping track of long lists of actions. To mitigate this, we decompose the problem by partitioning the actions applicable at the current state into $n$ sublists. The LLM is then queried iteratively with $n$ prompts derived from these partitions and returns $n$ index-action pairs. Each pair is, according to the LLM, the most likely action from its corresponding sublist to contribute to completing the task.

These action candidates form the action list for a final prompt, which selects the most preferable action from the $n$ index-action pairs. The agent executes this chosen action by applying the transition function $f_t$ to the current state, resulting in a new current state. The action is then added to the list of taken actions, and the process is repeated.

This process is presented in a simplified form in Figure~\ref{fig:diagram}, which only shows one step for action selection. In practice, as described above, this step is actually repeated $n$ times across the $n$ action list partitions and then once more across the $n$ chosen actions to select just one.

\paragraph{Error correction.}
\label{error_correction}
To account for unexpected output from the LLM, often referred to as hallucinations~\cite{hallucinatealways2024,hallucinate2023}, we use error correction triggered by mismatched index-action pairs $i,\,a$; that is, where the index and action do not align within the applicable action list. The specific method with which we correct for errors depends on the stage of selecting an action that the LLM is in.

When assembling the list of action candidates, a mismatched index-action pair will result in both the action at index $i$ of the applicable action list and any actions with the same VirtualHome language representation as $a$ being added to the candidate list. If the index or action cannot be determined, no action is added to the candidate list. 

When selecting the final action from the candidate list, if the index and action do not align within the applicable action list or the index and action cannot be determined, then the query is repeated. The LLM is informed in the new prompt that this is a repeated query, as well as what its previous response was.

This process is omitted from Figure~\ref{fig:diagram}, which only shows one step for action selection for the purpose of clarity. In practice, as described above, an LLM can be queried multiple times with the same prompt before the LLM is considered to have selected a valid action. 

\paragraph{Stopping Criteria.}
\label{stopping_criteria}
The system stops on one of four conditions. 
\begin{itemize}
    \item If the goal conditions are satisfied at the current state, the system will return the current sequence of actions taken as a successful solution to the task.
    \item If the failure conditions are satisfied at the current state, the system will terminate, having failed to find a successful solution.
    \item If the length of the current sequence of actions taken exceeds the limit on maximum solution length, the system will return the current sequence of actions taken and report that it has failed to find a solution within the solution length limits given.
    \item If the cumulative number of repeated queries exceeds the limit, the system will return the current sequence of actions taken and report that it has failed to find a solution within the repeated query limits given.
\end{itemize}
   
\section{Results}
\subsection{Experiment Setup}
Our system is tested on the 10 common household tasks used for evaluation by ProgPrompt~\cite{progprompt2023}. Each task has a high-level task description, low-level goal conditions and low-level failure conditions. All three are hand-written and known by the simulator. However, only the description is visible to the agent. All of our experiments use \texttt{gpt-4o-mini-2024-07-18} as the strategy-generating and action-selecting LLM, which is a scaled-down and substantially cheaper variant of the GPT-4~\cite{gpt4} architecture.

For all experiments, we set the limit on maximum solution length to 20, the limit on the cumulative number of repeated queries per repetition of an experiment to 10, and the partition size for the applicable action list to 100. The parameter which regulates the spread of the distribution from which the LLM output is sampled, known as the model temperature, is set to 0.2. Values closer to zero are known to result in less creative and varied outputs, which is preferable for generating accurate solutions in a low-level language with limited vocabulary. Each experiment is repeated 50 times.

\subsection{Main Experiments}
\begin{table*}[t]
\centering
\begin{tabular}{lccc|cccccccc}
\toprule
 & \multicolumn{3}{c}{Main Experiments} & \multicolumn{2}{c}{Original} & \multicolumn{2}{c}{Objects} & \multicolumn{2}{c}{Static} & \multicolumn{2}{c}{Dynamic}\\
 
\cmidrule(lr){1-1} \cmidrule(lr){2-4} \cmidrule(lr){5-6} \cmidrule(lr){7-8} \cmidrule(lr){9-10} \cmidrule(lr){11-12}

\textbf{Task Description} & \textbf{NG} & \textbf{LLG} & \textbf{HLG} & \textbf{G} & \textbf{G+S} & \textbf{G} & \textbf{G+S} & \textbf{G} & \textbf{G+S} & \textbf{G} & \textbf{G+S}\\

\cmidrule(lr){1-1} \cmidrule(lr){2-4} \cmidrule(lr){5-6} \cmidrule(lr){7-8} \cmidrule(lr){9-10} \cmidrule(lr){11-12}

\textit{watch tv} & 1.00 & 0.98 & 0.84 & 0.70 & 0.98 & 1.00 & 1.00 & 1.00 & 1.00 & 1.00 & 1.00\\
\textit{turn off light} & 1.00 & 1.00 & 1.00 & 0.0 & 1.00 & 0.38 & 1.00 & 1.00 & 1.00 & 1.00 & 1.00\\
\textit{throw away apple} & 0.02 & 0.02 & 0.02 & 0.0 & 0.02 & 0.0 & 0.0 & 0.0 & 0.02 & 0.0 & 0.02\\
\textit{make toast} & 0.02 & 1.00 & 0.48 & 0.0 & 1.00 & 0.0 & 0.22 & 0.0 & 0.90 & 0.02 & 0.86\\
\textit{eat chips on the sofa} & 0.22 & 1.00 & 1.00 & 1.00 & 1.00 & 1.00 & 1.00 & 1.00 & 1.00 & 1.00 & 1.00\\
\textit{put salmon in the fridge} & 0.90 & 1.00 & 0.98 & 0.78 & 1.00 & 0.02 & 0.98 & 0.44 & 1.00 & 0.38 & 1.00\\
\textit{brush teeth} & 0.12 & 0.10 & 0.10 & 0.0 & 0.10 & 0.0 & 0.24 & 0.0 & 0.08 & 0.0 & 0.46\\
\textit{wash the plate} & 0.0 & 0.02 & 0.08 & 0.0 & 0.02 & 0.0 & 0.14 & 0.0 & 0.24 & 0.0 & 0.18\\
\textit{microwave salmon} & 0.08 & 0.74 & 1.00 & 0.44 & 0.74 & 0.14 & 0.86 & 0.10 & 0.98 & 0.08 & 0.98\\
\textit{bring items* to the coffee table} & 0.70 & 1.00 & 0.98 & 0.0 & 1.00 & 0.0 & 0.24 & 0.08 & 0.52 & 0.08 & 0.40\\

\cmidrule(lr){1-1} \cmidrule(lr){2-4} \cmidrule(lr){5-6} \cmidrule(lr){7-8} \cmidrule(lr){9-10} \cmidrule(lr){11-12}

Task Averages & 0.41 & 0.69 & 0.65 & 0.29 & 0.69 & 0.25 & 0.57 & 0.36 & 0.67 & 0.36 & 0.69\\
\bottomrule
\end{tabular}
\caption{Evaluation on 10 household tasks. We evaluate our method on three types of guide heuristics: none (NG), low-level (LLG) and high-level (HLG). Also compared are low-level guides alone (G) and guide-assisted local search (G+S), across four levels of environment information: no information (Original), list of objects (Objects), list of objects with static properties (Static), and list of objects with static and dynamic properties (Dynamic). *items are `coffeepot and cupcake'}
\label{tab:main_table}
\end{table*}
The success rates for our experiments, that is, the proportion of output solutions that achieve all associated goal conditions for a task, are presented in Table~\ref{tab:main_table}. The tasks are ordered in ascending order by the minimum length of a plan required to solve them.

In order to evaluate the effects of utilising a guide as a heuristic, each experiment is a combination of a task description and either a high-level guide (HLG), a low-level guide (LLG) or no guide (NG), with which to aid the LLM-driven local search. 

As our low-level LLM-generated guides function as stand-alone plans for their respective tasks, we are also able to evaluate the effect of the subsequent local search on task success rates. We do this by comparing the success rate of using the generated low-level guide (G) as a solution to the success rate of our LLM-driven local search, which uses that guide as a heuristic (G + S).

The guide-generating LLM has less information on environment dynamics and objects within our environment compared to the action-selecting LLM, which has access to the semantic information contained within the lists of applicable actions. As such, we conduct our comparison between low-level guides and guide-assisted local search across four levels of environment knowledge. When conducting our main experiments comparing guide types (NG, LLG, HLG), the guide prompt contained no information about the environment. We extend this by testing guide generation with progressively detailed prompts: 1) containing a list of all objects in the environment; 2) containing a list of objects along with their static properties (e.g., \texttt{CAN OPEN, HAS SWITCH, GRABBABLE}); and 3) containing objects, their static properties, and their dynamic properties (e.g., \texttt{OPEN/CLOSED, ON/OFF}).

\paragraph{Outcome.}
We conclude that our method is effective at generating plans for solving simple natural language tasks in a virtual environment, even when only leveraging the world knowledge of the LLM at the point of action selection; that is, local search with no guide. Our method's performance is improved with the addition of guides as heuristics, with a slight advantage in performance for low-level guides over high-level ones. When enhanced with a low-level guide, our LLM-aided local search achieves a success rate of 69 percent on average, 22 percentage points above the best-performing ProgPrompt setup, and 32 percentage points above ProgPrompt running on GPT-4 architecture~\cite{progprompt2023}. Without a low-level guide, our method (0.41) performs slightly worse than ProgPrompt (0.47).

The results from comparing just the guides to local search using the guides as a heuristic show that local search has a significant impact on the success rate of our method. Across all four levels of information, local search (G+S) outperformed just the guide (G) by at least 30 percentage points. Notably, an increase in the amount of information provided to the guide-generating LLM does not close the gap between G and G+S. 

The results show that guides generated with no environment information do a better job of guiding local search (0.69) than guides generated with a list of objects within the environment (0.57). This is roughly proportional to the decrease in the performance of the guides themselves, from 29 percent to 25 percent. The majority of this effect is due to the poor performance of local search with object-aware guides on the `make toast' and `bring the coffeepot and the cupcake to the coffeetable' tasks.

\paragraph{Adversarial Environments.} We extend our experiments on the gap between guides and guide-assisted local search to an `adversarial' environment where the variable properties of key objects may differ from the LLM's world model. In this environment the tv is \texttt{ON}, and the bin, fridge and microwave are \texttt{OPEN}.  Our results are presented in Table~\ref{tab:main_adversarial_guide_evaluation}

In an adversarial environment, the impact of local search is even larger than in our original environment. When used directly as a solution, guides are unable to solve a single task, with the exception of guides with information on dynamic properties for `put salmon in the fridge'. This is an interesting result, as our adversarial changes are visible to the guide-generating LLM at the `dynamic properties' level of information. However, even at that level, three of the four tasks were not solved a single time. Our results show that, while the overall performance of the guide-assisted local search in an adversarial environment decreases for three of the four levels of information, it is much more resilient to the change than the guides on their own. 

\begin{table*}[t]
\centering
\begin{tabular}{lcccccccc}
\toprule
 & \multicolumn{2}{c}{Original} & \multicolumn{2}{c}{Objects} & \multicolumn{2}{c}{Static} & \multicolumn{2}{c}{Dynamic}\\
\cmidrule(lr){1-1} \cmidrule(lr){2-3} \cmidrule(lr){4-5} \cmidrule(lr){6-7} \cmidrule(lr){8-9}
\textbf{Task} & \textbf{G} & \textbf{G\texttt{+}S} & \textbf{G} & \textbf{G\texttt{+}S} & \textbf{G} & \textbf{G\texttt{+}S} & \textbf{G} & \textbf{G\texttt{+}S}\\
\cmidrule(lr){1-1} \cmidrule(lr){2-3} \cmidrule(lr){4-5} \cmidrule(lr){6-7} \cmidrule(lr){8-9}
\textit{watch tv} & 0.0 & 0.64 & 0.0 & 0.82 & 0.0 & 0.52 & 0.0 & 0.62\\
\textit{throw away apple} & 0.0 & 0.10 & 0.0 & 0.0 & 0.0 & 0.14 & 0.0 & 0.54\\
\textit{put salmon in the fridge} & 0.0 & 1.00 & 0.0 & 1.00 & 0.0 & 1.00 & 0.52 & 1.00\\
\textit{microwave salmon} & 0.0 & 0.88 & 0.0 & 0.50 & 0.0 & 0.88 & 0.0 & 0.90\\
\cmidrule(lr){1-1} \cmidrule(lr){2-3} \cmidrule(lr){4-5} \cmidrule(lr){6-7} \cmidrule(lr){8-9}
Adversarial Task Averages & 0.00 & 0.66 & 0.00 & 0.58 & 0.00 & 0.64 & 0.13 & 0.77\\
Non-Adversarial Task Averages & 0.48 & 0.69 & 0.29 & 0.71 & 0.39 & 0.75 & 0.36 & 0.75\\
\bottomrule
\end{tabular}
\caption{A subset of four Table~\ref{tab:main_table} tasks in an `adversarial' environment. The tv is \texttt{ON}, and the bin, fridge and microwave are \texttt{OPEN}. The adversarial averages for these tasks are compared against their non-adversarial averages (from Table~\ref{tab:main_table}).}
\label{tab:main_adversarial_guide_evaluation}
\end{table*}

\paragraph{Executability}
Selecting actions with local search sidesteps the difficulties with executability faced by similar research. As action selection is always performed on a list of applicable actions, the final solution is always perfectly executable. Furthermore, as our approach operates directly in the VirtualHome language, we do not require a potentially error-prone translation step. 

\subsection{Error Analysis}
Table~\ref{tab:hint_evaluation} shows an evaluation of five modified experiments for three tasks which were particularly low-scoring in our main experiments from Table~\ref{tab:main_table}. Our findings suggest that errors can often occur due to discrepancies between the LLM's `envisioned' environment and the actual environment it finds itself in. This may be due to the agent having no access to the formal action schema within its environment, to the internal state representations used by the simulator, or to the internal goal conditions for a given task.

\begin{table*}[t]
\centering
\begin{tabular}{lcccccc}
\toprule
& \multicolumn{3}{c}{Original} & \multicolumn{3}{c}{Modified}\\
\cmidrule(lr){1-1} \cmidrule(lr){2-4} \cmidrule(lr){5-7}
\textbf{Experiment Variation} & \textbf{NG} & \textbf{LLG} & \textbf{HLG} & \textbf{NG} & \textbf{LLG} & \textbf{HLG}\\
\cmidrule(lr){1-1} \cmidrule(lr){2-4} \cmidrule(lr){5-7}
\textit{`throw away apple. you must put the apple in the bin'} & 0.02 & 0.02 & 0.02 & 0.40 & 0.88 & 1.00\\
\textit{`throw away apple'}. All \texttt{DROP} actions are removed. & 0.02 & 0.02 & 0.02 & 0.68 & 1.00 & 1.00\\
\textit{`wash the plate. don't forget to turn on the faucet!'} & 0.0 & 0.02 & 0.08 & 0.68 & 0.84 & 0.56\\
\textit{`brush teeth. you will need a hand free to open the toothpaste.'} & 0.12 & 0.10 & 0.10 & 0.28 & 0.34 & 0.66\\
\textit{`brush teeth'}. The toothpaste is always \texttt{OPEN}. & 0.12 & 0.10 & 0.10 & 1.00 & 0.90 & 0.90\\
\bottomrule
\end{tabular}
\caption{An evaluation on modifications of three low-performing tasks from our main experiments in Table~\ref{tab:main_table}. The success rates of the modified tasks (Modified) are compared against their original success rates (Original) across three types of guide heuristic (NG, LLG, HLG).}
\label{tab:hint_evaluation}
\end{table*}

\paragraph{Throw Away Apple.}
When tasked with throwing away the apple, the LLM will most often direct the agent to \texttt{PICKUP} the apple, \texttt{OPEN} the bin and then \texttt{DROP} the apple. In a real-world environment, `grab the apple and drop it into the bin' is not an unreasonable answer to `how do I throw away an apple?'. However, in a strict, STRIPS-based environment, the effects of the \texttt{DROP} action are simple and exact. In our domain, it removes the \texttt{HOLDS RH} or \texttt{HOLDS LH} relation between the agent and the object (the object is no longer in the agent's hand). This is insufficient to satisfy the goal conditions for the task, which specify that there must be an \texttt{INSIDE} relation between the apple and the bin. In our domain, this relation is only achievable through the \texttt{PUTIN} action. Adding a small hint to the prompt (`you must put the apple in the bin') has a substantial impact on the success rate of the task. Even more substantial is simply removing the \texttt{DROP} action schema entirely. This suggests that the issue was not with the agent's understanding that the apple should end up in the bin but rather with its ability to understand the dynamics and limitations of its environment.

\paragraph{Wash the Plate.}
Washing the plate is a rather ambiguous task. For our experiments, the goal conditions are realised through a \texttt{WASH} action, which is applicable when the character is holding an object next to a sink object with the \texttt{FILLED} property. A sink object can acquire this property when its corresponding faucet object is switched on via a \texttt{SWITCHON} action. It is important to note that a strong case could be made for several other interpretations, so this is only one of many valid ways of defining a set of goal conditions for the task. However, the key issue for our agent is not the vague nature of the task description but rather the ambiguous way in which you must interact with the sink, which is a separate object from its faucet. We demonstrate this by adding a minor hint to `turn on the faucet' to the original task description, which brings the success rate to above 50 percent.

\paragraph{Brush Teeth.}
The agent can brush its teeth by pouring toothpaste onto the toothbrush and then using it with the \texttt{USE} action. In our environment, the toothpaste is \texttt{CLOSED} and can only be opened with one free hand while being held with the other. This is a point of difficulty for the agent, which will usually attempt to pick up both the toothbrush and the toothpaste before trying to \texttt{OPEN} the toothpaste. This is not an unreasonable solution, as an adult human would usually be able to open a tube of toothpaste even while holding a toothbrush. However, in our environment, it will find no applicable \texttt{[OPEN]<toothpaste>} action and often subsequently fail to solve the task, leading to a 10 percent success rate. We note that, when the toothpaste is always \texttt{OPEN}, this is a simple task for the agent, with close to perfect success rates.  

Hinting that `you will need a hand free to open the toothpaste' does not bring the success rate to the same levels as keeping the toothpaste \texttt{OPEN}. This suggests that this issue may not be due to a misunderstanding of environment dynamics but rather due to the LLM failing to solve the logic sub-puzzle of opening the toothpaste with limited hands. 

\section{Conclusions}
We present an approach for leveraging the internal world knowledge and reasoning capabilities of LLMs to inform action selection when using local search. Our approach is capable of generating plans for common household tasks, and can be extended by utilising solution estimates as a further heuristic to guide the local search.  

We show that LLM-aided local search is capable of solving simple tasks in a household environment with a comparable success rate to similar methods. Our solutions are fully executable and are generated directly in the VirtualHome language, demonstrating that strong results can be achieved without an intermediate language or a translation step. 

Furthermore, we show that enhancing the search with a solution estimate as a heuristic results in better performance than local search or the solution estimate on their own, regardless of whether the estimate is represented in a low-level or a high-level language. This increase in performance places the success rate of our approach at 22 percent higher than that of ProgPrompt on the same set of household tasks.   

We further give clear evidence that the impact of local search cannot be replicated by providing more environment information to the LLM tasked with generating solution estimates. This impact is even more substantial in an `adversarial' environment, which differs from the assumed world model of the LLM. This demonstrates the comparative resilience of local search to unexpected variables within its environment.

\paragraph{Future Work.}
Our approach leverages the LLM's internal model of a standard household environment. Lifting such an environment into a finite language representation often causes discrepancies between the LLM environment model and the environment in which the agent operates. Future work could explore methods to pass a formal description of environment rules to the LLM as part of the task input through techniques such as RAG~\cite{lewis:rag}.

Our approach will likely have difficulty with tasks that require an increased amount of reasoning skills. The tasks used in our paper mostly represent variations in moving objects around the house. The LLM is only queried for a single action at a time, which makes big-picture reasoning, such as opening toothpaste with finite hands, more difficult. It could be valuable to leverage decomposition techniques such as Chain of Thought~\cite{chainofthought2022} throughout parts of the search to improve performance.

Local search can get stuck in local maxima or through taking irreversible actions, which makes backtracking impossible. This could be mitigated with an algorithm like beam search~\cite{aima2016}, which would also have the potential to increase success rates by allowing the algorithm to pick and choose from a variety of plans.

\section{Declarations}
\label{section:declaration}
To enhance the clarity of our research and promote future work, we will open-source our code and data at \url{https://github.com/andrey-borro/llm-common-sense}.

\bibliographystyle{named}
\bibliography{references}

\clearpage
\onecolumn
\appendix

\section{Task Goal Conditions}

\section{LLM Prompts}
\subsection{Prompt to generate a high-level guide with no environment information}
\begin{lstlisting}
You are a system in charge of creating hypothetical plans in a simulated environment for a robot to carry out. The robot has a finite number of commands that it understands. 
The following is a list of actions available to you in this simulated environment:
walk (1), close (1), cut (1), drink (1), drop (1), eat (1), grab (1), greet (1), lie (1), move (1), open (1), plugin (1), plugout (1), pour (2), placeon (2), putin (2), puton (1), read (1), sit (1), sleep (0), standup (0), switchoff (1), switchon (1), takeoff (1), touch (1), type (1), use (1), wakeup (0), wash (1), watch (1), wipe (2)
The walk action can usually take you directly to any item in the house. Unless you are holding it, it is safe to assume that you should walk to an item before interacting with it.
The VERB is the action name and the number in brackets is how many objects it applies to, e.g. 'standup (0) = Stand Up', 'lookat (1) = Look at the picture', 'putin (2) = Put the cake in the oven'.

[ENVIRONMENT INFO WOULD GO HERE]

I want you to give me the proposed plan as an series of actions, with no comments or indexing.

Here is an example of a possible output to  "What is a high-level plan for 'pack your pencilcase'?".

Walk to the pencilcase
Grab the pencilcase
Walk to the schoolbag
Put the pencilcase in the schoolbag

Here is another example for a possible output to "What is a high-level plan for 'put two spoons and a knife into the dishwasher'?"
 
Walk to the dishwasher
Open the dishwasher
Walk to the spoon
Grab the spoon
Walk to the dishwasher
Put the spoon in the dishwasher
Walk to another spoon
Grab the spoon
Walk to the dishwasher
Put the spoon in the dishwasher
Walk to the knife
Pick up the knife
Walk to the dishwasher
Put the knife in the dishwasher


What is a high-level plan for 'put the cat in the bathtub'?
\end{lstlisting}

\subsection{Environment information with dynamic properties}
\begin{lstlisting}
bathroom: properties - {ROOM} | states - {}
floor: properties - {} | states - {}
wall: properties - {} | states - {}
ceiling: properties - {} | states - {}
rug: properties - {SURFACES} | states - {}
curtains: properties - {CAN_OPEN, COVER_OBJECT} | states - {CLOSED}
ceilinglamp: properties - {} | states - {}
walllamp: properties - {} | states - {}
bathtub: properties - {LIEABLE} | states - {}
towelrack: properties - {} | states - {}
wallshelf: properties - {} | states - {}
stall: properties - {} | states - {}
toilet: properties - {SITTABLE, CAN_OPEN, CONTAINERS} | states - {OFF, CLOSED}
bathroomcabinet: properties - {} | states - {CLOSED}
bathroomcounter: properties - {SURFACES} | states - {CLOSED}
sink: properties - {RECIPIENT, CONTAINERS} | states - {}
faucet: properties - {HAS_SWITCH} | states - {OFF}
door: properties - {CAN_OPEN} | states - {OPEN}
doorjamb: properties - {} | states - {}
towel: properties - {GRABBABLE, COVER_OBJECT} | states - {}
perfume: properties - {} | states - {}
deodorant: properties - {} | states - {}
hairproduct: properties - {GRABBABLE, POURABLE} | states - {}
facecream: properties - {GRABBABLE, POURABLE, CAN_OPEN, CREAM} | states - {CLOSED}
...
\end{lstlisting}

\subsection{System message for action-selection prompt}
\begin{lstlisting}
You are the decision system of a robot excecuting tasks in a simulated environment. 
You will receive: 

(a) A high-level goal/task in natural language (e.g. watch TV)
(b) A pre-generated estimate of what the solution may look like
(c) A list of the actions you have already taken towards satisfying your task
(d) A list of the actions available for you to pick from for the next step.

You will return the next action that the robot should do to complete the task, in the form {INDEX [VERB]<item_1><item_2>...<item_n>}. Precede this with a very brief explanation of your reasoning. Sometimes, none of the actions will seem particularly good. It is enough to pick the one you think is most likely. 

All available actions will be of the form INDEX [VERB]<item_1><item_2>...<item_n>, with most having one or two items. I want all outputs to only be a singular low-level action. 
The INDEX number is only there for the robots own internal use, you should not take it into consideration, but it should be part of the output. Take care not to return an action that is not present in the action list provided.  

Keep in mind that in this environment, you must usually WALK to an object before you can interact with it, unless you have picked it up and are holding it. For example, if you want to [GRAB]<onion> but it isnt in the list, it may be worth checking if [WALK]<onion> is. Walking to any object will take you away from all other objects, so you must walk back to them if you want to interact with them again.

Watch out for chains of repeated or redundant actions. These could signify that a new approach is necessary.

In order to assist you, a separate system has generated its estimate for what a full high-level plan could look like (input item b). It will most likely not be fully accurate, and have missing, additional or impossible actions, but can be useful in guiding your decisions. 

Occasionally, the prompt will let you know that it is a repeated query. This means that the answer given previously was not one of the available actions, and you need to reselect one that is. Pay special attention to the index - if one off the available actions is `12 [WALK]<bottle>`, `15 [WALK]<bottle>` will not be a valid output.

For example, an input will look like:
'''
'what is the first step to high-level goal x', or 'what is the next step to high-level goal x'.

**A pregenerated high-level plan estimate (might have missing, additional or impossible actions) has been provided:**
Step 1 
Step 2
...
Step N

**The steps you have already taken towards this goal are:**\\
VERB<item>
VERB<item>
...
VERB<item><item>

**The list of actions available to you are:**
INDEX VERB<item>
INDEX VERB<item>
INDEX VERB<item><item>
INDEX VERB<item>
'''

Your output would look like:
*Reasoning here*

{INDEX VERB<action>}
\end{lstlisting}

\twocolumn
\subsection{Example of action-selection prompt [actions 0-99]}
\begin{lstlisting}
What is the next step to high-level goal 'microwave salmon'?
**A pregenerated low-level plan estimate (might have missing, additional or impossible actions) has been provided:**
walk | microwave  
open | microwave  
walk | salmon  
grab | salmon  
walk | microwave  
putin | salmon | microwave  
close | microwave  
switchoff | microwave  

**The actions you have already taken are**:
[WALK]<microwave>
[OPEN]<microwave>

**The list of actions available to you are:**
0 [WALK]<bathroom>
1 [WALK]<floor>
2 [WALK]<floor>
3 [WALK]<floor>
4 [WALK]<floor>
5 [WALK]<floor>
6 [WALK]<floor>
7 [WALK]<floor>
8 [WALK]<wall>
9 [WALK]<wall>
10 [WALK]<wall>
11 [WALK]<wall>
12 [WALK]<wall>
13 [WALK]<wall>
14 [WALK]<wall>
15 [WALK]<ceiling>
16 [WALK]<ceiling>
17 [WALK]<ceiling>
18 [WALK]<ceiling>
19 [WALK]<ceiling>
20 [WALK]<ceiling>
21 [WALK]<rug>
22 [WALK]<curtains>
23 [WALK]<curtains>
24 [WALK]<curtains>
25 [WALK]<ceilinglamp>
26 [WALK]<walllamp>
27 [WALK]<walllamp>
28 [WALK]<walllamp>
29 [WALK]<bathtub>
30 [WALK]<towelrack>
31 [WALK]<towelrack>
32 [WALK]<towelrack>
33 [WALK]<towelrack>
34 [WALK]<wallshelf>
35 [WALK]<stall>
36 [WALK]<toilet>
37 [WALK]<stall>
38 [WALK]<curtains>
39 [WALK]<bathroomcabinet>
40 [WALK]<bathroomcounter>
41 [WALK]<sink>
42 [WALK]<faucet>
43 [WALK]<door>
44 [WALK]<doorjamb>
45 [WALK]<towel>
46 [WALK]<towel>
47 [WALK]<towel>
48 [WALK]<perfume>
49 [WALK]<deodorant>
50 [WALK]<hairproduct>
51 [WALK]<hairproduct>
52 [WALK]<facecream>
53 [WALK]<plate>
54 [WALK]<toothpaste>
55 [WALK]<painkillers>
56 [WALK]<waterglass>
57 [WALK]<toothbrush>
58 [WALK]<barsoap>
59 [WALK]<towel>
60 [WALK]<towel>
61 [WALK]<candle>
62 [WALK]<window>
63 [WALK]<lightswitch>
64 [WALK]<character>
65 [WALK]<washingmachine>
66 [WALK]<bedroom>
67 [WALK]<floor>
68 [WALK]<floor>
69 [WALK]<floor>
70 [WALK]<floor>
71 [WALK]<floor>
72 [WALK]<floor>
73 [WALK]<floor>
74 [WALK]<floor>
75 [WALK]<floor>
76 [WALK]<floor>
77 [WALK]<wall>
78 [WALK]<wall>
79 [WALK]<wall>
80 [WALK]<wall>
81 [WALK]<wall>
82 [WALK]<wall>
83 [WALK]<wall>
84 [WALK]<wall>
85 [WALK]<window>
86 [WALK]<ceiling>
87 [WALK]<ceiling>
88 [WALK]<ceiling>
89 [WALK]<ceiling>
90 [WALK]<ceiling>
91 [WALK]<ceiling>
92 [WALK]<ceiling>
93 [WALK]<ceiling>
94 [WALK]<ceiling>
95 [WALK]<ceilinglamp>
96 [WALK]<tablelamp>
97 [WALK]<tablelamp>
98 [WALK]<garbagecan>
99 [WALK]<nightstand>
\end{lstlisting}

\onecolumn
\subsection{Example of action-selection prompt [candidate actions]}
\begin{lstlisting}
What is the next step to high-level goal 'microwave salmon'?
**A pregenerated low-level plan estimate (might have missing, additional or impossible actions) has been provided:**
walk | microwave  
open | microwave  
walk | salmon  
grab | salmon  
walk | microwave  
putin | salmon | microwave  
close | microwave  
switchoff | microwave  

**The actions you have already taken are**:
[WALK]<microwave>
[OPEN]<microwave>

**The list of actions available to you are:**
3 [WALK]<floor>
100 [WALK]<bookshelf>
296 [WALK]<microwave>
310 [WALK]<salmon>
400 [WALK]<rug>
\end{lstlisting}

\subsection{Error correction prompt prefix}
\begin{lstlisting}
Error Correction prompt format]
This query is being repeated for you. Your previous response was [PREVIOUS RESPONSE], but this was not one of the options available to you. Please select one of the available actions and answer the prompt in the correct format. 

[ORIGINAL PROMPT GOES HERE]
\end{lstlisting}

\end{document}